\documentclass[runningheads]{llncs}
\usepackage{graphicx}
\usepackage[table]{xcolor}
\usepackage{tikz}
\usepackage{comment}
\usepackage{amsmath,amssymb} 
\usepackage{bbm, dsfont}
\usepackage{algorithm}
\usepackage{algorithmic}
\usepackage{bbding}
\usepackage{cite}
\usepackage{url}
\usepackage{csquotes}
\usepackage{subcaption}
\usepackage{multirow}
\usepackage{booktabs}
\usepackage{threeparttable}

\usepackage{newfloat}
\usepackage{listings}
\usepackage{nccmath}
\usepackage[accsupp]{axessibility}  

\usepackage[pagebackref,breaklinks,colorlinks]{hyperref}
\usepackage[capitalize]{cleveref}
\crefname{section}{Sec.}{Secs.}
\Crefname{section}{Section}{Sections}
\Crefname{table}{Table}{Tables}
\crefname{table}{Tab.}{Tabs.}
\usepackage[width=122mm,left=12mm,paperwidth=146mm,height=193mm,top=12mm,paperheight=217mm]{geometry}

\begin{document}

\pagestyle{headings}
	\mainmatter
	
	\title{Learning Unbiased Transferability for Domain Adaptation by Uncertainty Modeling} 

	\titlerunning{Learning Unbiased Transferability for DA by Uncertainty Modeling}
	\author{Jian Hu\inst{1}\inst{\star}, Haowen  Zhong\inst{2}\thanks{Equal contribution}, Fei Yang\inst{2}, \\Shaogang Gong\inst{1},  Guile Wu\inst{},  Junchi Yan\inst{3}}
	\authorrunning{J. Hu et al.}
	%
	\institute{Queen Mary University of London \and Zhejiang Lab \and Shanghai Jiao Tong University \\
	\email{\{jian.hu,s.gong\}@qmul.ac.uk}, \email{guile.wu@outlook.com}\\\email{\{zhonghw,yangf\}@zhejianglab.com}, \email{yanjunchi@sjtu.edu.cn}}
	\maketitle

\begin{abstract}
    Domain adaptation (DA) aims to transfer knowledge learned from a labeled source domain to an unlabeled or a less labeled but related target domain. Ideally, the source and target distributions should be aligned to each other equally to achieve unbiased knowledge transfer. However, due to the significant imbalance between the amount of annotated data in the source and target domains,
    usually only the target distribution is aligned to the source domain, leading to adapting unnecessary source specific knowledge to the target domain, i.e., biased domain adaptation. To resolve this problem, in this work, we delve into the transferability estimation problem in domain adaptation and propose a non-intrusive Unbiased Transferability Estimation Plug-in (UTEP) by modeling the uncertainty of a discriminator in adversarial-based DA methods to optimize unbiased transfer. We theoretically analyze the effectiveness of the proposed approach to unbiased transferability learning in DA. Furthermore, to alleviate the impact of imbalanced annotated data, we utilize the estimated uncertainty for pseudo label selection of unlabeled samples in the target domain, which helps achieve better marginal and conditional distribution alignments between domains. Extensive experimental results on a high variety of DA benchmark datasets show that the proposed approach can be readily incorporated into various adversarial-based DA methods, achieving state-of-the-art performance.
\end{abstract}

\section{Introduction}
With the rise of deep neural networks,  convolutional neural networks (CNNs)~\cite{iandola2016squeezenet} and transformers~\cite{parmar2018image},
supervised learning based on deep neural networks has shown promising performance in various vision and language tasks.
Conventional supervised learning methods mostly assume that the training (source) domain and the test (target) domain
are subject to the i.i.d. (independent and identically distributed) hypothesis.
As a consequence, they usually show poor generalization in the test domain in the existence of a domain gap~\cite{Hoffman18}.
Moreover, deep neural networks have ravenous appetite for a large amount of labeled data,
which explores extra linkage between labelled data in a source domain and unlabeled data in a target domain.
Domain adaption aims to resolve this quandary by leveraging previously labeled datasets~\cite{Wang19,Pan19} to realize effective knowledge transfer across domains.
DA can further divide into unsupervised domain adaptation (UDA) and semi-supervised domain adaptation (SSDA) based on the accessibility of labeled data in the target domain.

\begin{figure}[tb!]
    \centering
    \includegraphics[scale=0.43]{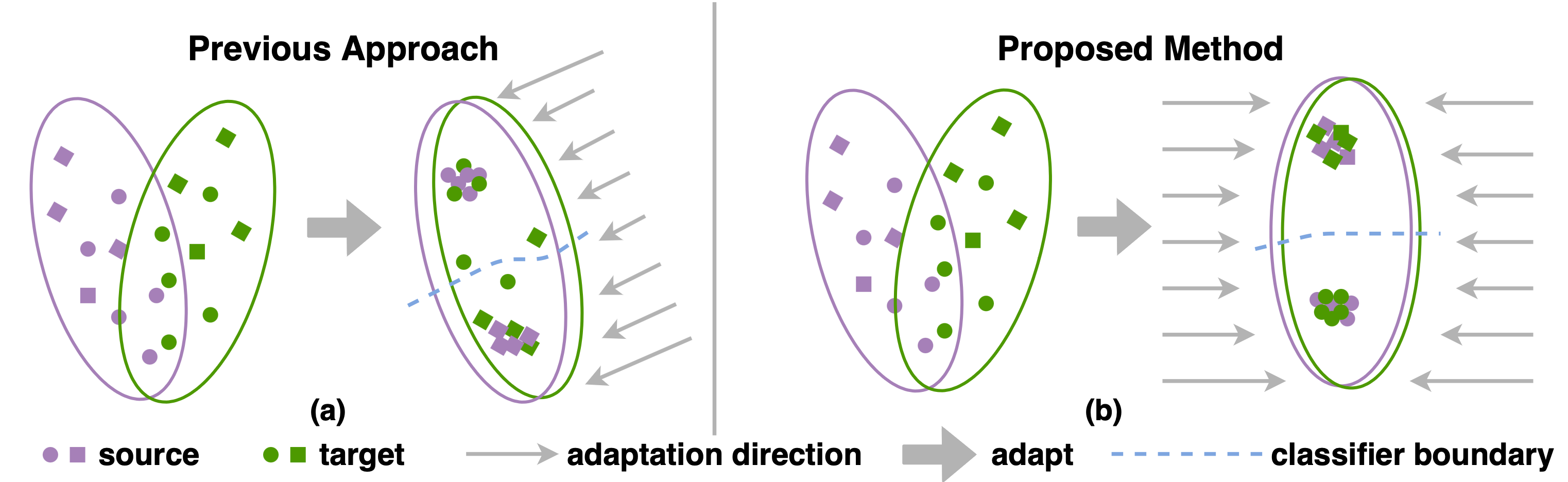}
    \caption{
    \textbf{Motivation for Unbiased Transferability Estimation.}
   The left and right parts in (\textbf{a}) (\textbf{b}) are distributions of source and target domains before and after adaptation respectively.
    (\textbf{a}) Previous methods~\cite{Ganin16,Long16} focus on aligning target samples to source well-clustered classes (distribution), leading to misalignment near the classifier boundary and transferring unnecessary source specific knowledge to the target, and the decision boundaries remain ambiguous for some target samples near the boundaries.
    (\textbf{{b}}) Our method enforces source and target distributions to align to each other equally, to help unbiased DA to achieve better marginal and conditional distribution alignment.
    The target decision boundary is further optimized with uncertainty-based pseudo label selection.}
    \label{fig:idea_illustration}
\end{figure}
Transferability indicates the ability of representations to bridge the discrepancy across domains\cite{chen2019transferability}.
Contemporary deep DA methods mainly focus on exploring cross-domain transferability to narrow the gap between domains~\cite{Saenko10,Yosinski14}.
There are two typical approaches:
1) High-order moment matching~\cite{Long15,inproceedings}, which reduces the distribution discrepancy between domains by
minimizing the distance between high-dimensional features;
2) Adversarial learning~\cite{Ganin16,Jian1,Tzeng17},
in which a domain discriminator and a feature extractor play a two-player game to align distributions between domains.
%
Although existing DA methods have shown promising performance on resolving domain shift, there remain two major problems.
First, most existing methods tend to promote the alignment of domains to encourage transferability,
which is evaluated by measuring the distribution discrepancy (estimated transferability)~\cite{0DOI,2017Conditional}.
However, even if the distribution discrepancy is completely eliminated,
the deviation of the estimated and real transferability still exists, leading to negative transfer (see Figure~\ref{fig:idea_illustration}(a)).
Second, existing approaches only consider aligning the overall distributions between domains
and tend to assign larger weights to samples that are not sufficiently transferred~\cite{2017Conditional,wen2019bayesian}.
However, since the number of annotations in the source and target domains is extremely imbalanced,
the transferability is usually estimated by the target sample score on the source domain classifier,
resulting in a strong bias towards the source domain (see Figure~\ref{fig:idea_illustration}(a)).
In other words, under the supervision of source labeled samples,
the source domain distribution is well learned and hardly affected by the target one,
while the target 
domain distribution tends to match the source domain instead of encouraging the source to align the target domain.
As a consequence, the target domain is not only learning the domain invariant knowledge (features)
but also learning domain specific knowledge of the source domain,
resulting in \emph{biased domain adaptation}.
Therefore, to address these two problems, it is essential to learn
unbiased transferability to facilitate knowledge transfer across domains in domain adaptation.

\begin{table}[tb!]
\small
 \centering
 \caption{A comparison of uncertainty modeling methods in DA.}
    \resizebox{0.6\columnwidth}{!}{
    \begin{tabular}{|c|c|c|c|c|}
    \hline
        {\multirow{2}{*}{Methods}} & online & non-intrusive & unbiased & uncertainty \\ 
         & learning   & design & transfer& usage \\ \hline
        {CPCS\cite{2020Calibrated}} & \Checkmark   & \XSolidBrush  & \XSolidBrush  & weighting for calibration\\ \hline
        {PACET\cite{liang2019exploring}} &\Checkmark   & \XSolidBrush  & \XSolidBrush & sample selection \\ \hline
        {BUM\cite{wen2019bayesian}} &\Checkmark  & \XSolidBrush  & \XSolidBrush & weighting\\ \hline
        {TransCal\cite{2020Transferable}} &\XSolidBrush  & \Checkmark  & \XSolidBrush  & post-hoc calibration \\ \hline
        {CADA\cite{kurmi2019attending}} &\Checkmark  & \XSolidBrush  & \XSolidBrush  & weighting for attention \\ \hline
        {\multirow{2}{*}{UTEP(Ours)}} &{\multirow{2}{*}{\Checkmark}}   & {\multirow{2}{*}{\Checkmark}}  & {\multirow{2}{*}{\Checkmark}} &         weighting + regularization\\
        &&&&+ sample selection\\\hline
    \end{tabular}}
    \centering

    \label{tab:compare}
\end{table}

\begin{figure*}[tb!]
  \centerline{\includegraphics[width=1.0\textwidth] {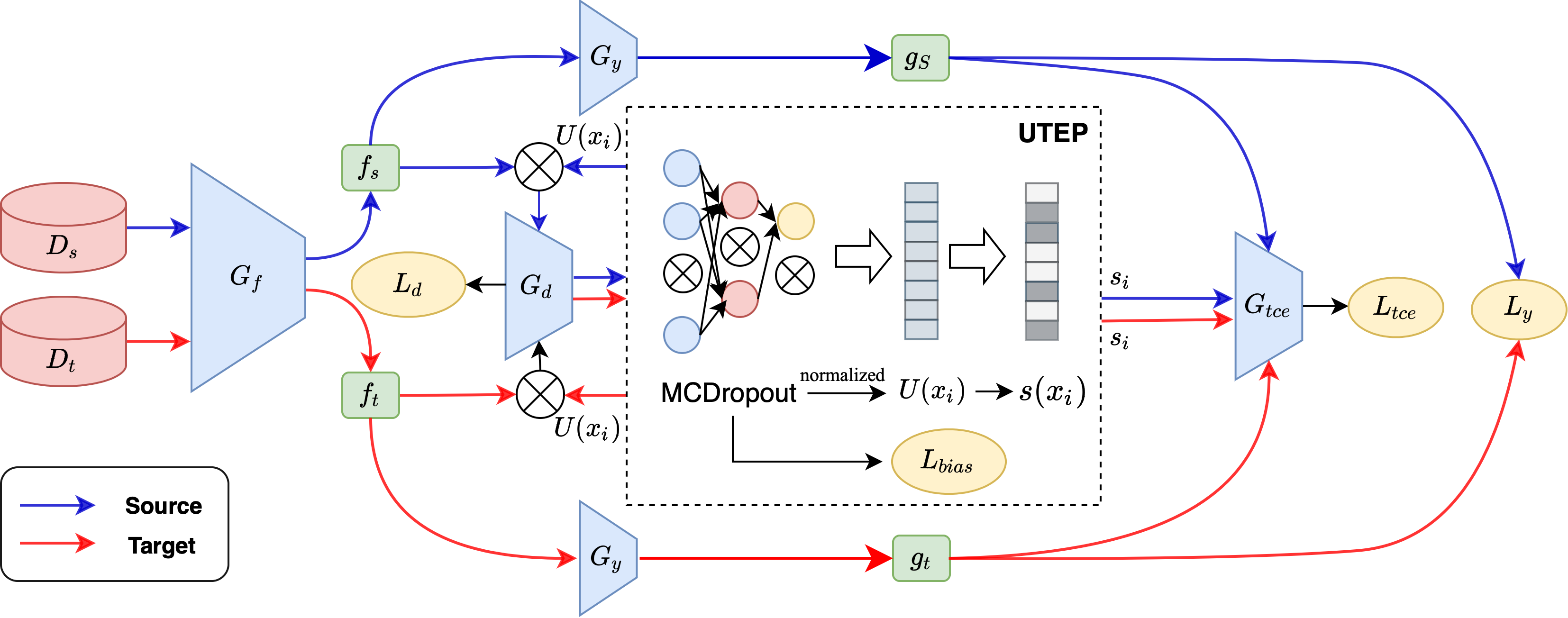}}
  \caption{The proposed UTEP framework.
  $G_f$ is the feature extractor. $G_d$ is the domain discriminator. $G_y$ is the source classifier. $G_{tce}$ is the pseudo label cluster. $f_s$ and $f_t$ are feature representations defined by $f=G_f(x)$, while $g_s$ and $g_t$ are the classifier prediction obtained from $G_y$. UTEP is the unbiased transferability estimation plug-in.
  This module can not only evaluate the unbiased transferability by minimizing $\mathcal{L}_{bias}$ and adding weight $\mu(x)$ on the discriminator with discriminator loss $L_d$, but also can generate $s(x)$ to evaluate the reliability of pseudo labels to facilitate learning with $L_{tce}$ and $L_y$.}
  \label{fig:architecture}
\end{figure*}

In this work, we focus on the transferability problem in domain adaptation
and present a non-intrusive Unbiased Transferability Estimation Plug-in (UTEP) model by uncertainty variance modeling.
Specifically, inspired by~\cite{2020Transferable} which uses a post-hoc strategy to re-calibrate existing DA methods,
we theoretically analyze the transferability of adversarial-based DA methods
and propose to model the uncertainty variance of a discriminator to minimize biased transfer between domains (see Figure~\ref{fig:idea_illustration}(b)).
Different from~\cite{2020Transferable}, UTEP incorporates unbiased transferability estimation into an online model learning process rather than post-processing,
which helps mitigate negative transfer and achieve better alignment between domains for unbiased domain adaptation.
Meanwhile, as the estimated uncertainty also reveals the reliability of unlabeled samples from the target domain,
we use it for pseudo label selection to achieve better marginal and conditional distribution alignment between source and target domains.
The proposed UTEP is plug-and-play and can be easily incorporated into various adversarial based DA methods.
An overview of UTEP is depicted in Figure~\ref{fig:architecture}. The \textbf{contributions} of this work are in three-fold:

\noindent 1) To address the transferability bias due to the imbalance between the quantity of labeled data from the source and target domains, unlike prior works\cite{2020Calibrated,2020Transferable} that model the classifier to obtain better calibrated prediction, which still has a strong bias toward source domain, we propose a novel non-intrusive Unbiased Transferability Estimation Plug-in (UTEP) model for DA by uncertainty variance modeling of a discriminator. 2) To quantify and minimize the bias of transferability, we theoretically analyze the cause of the bias and show that our method can alleviate it by lowering its upper bound. To our best knowledge, this is the first work to explore the unbiased knowledge transfer rather than model calibration via uncertainty modeling in domain adaptation. 3) We plug our technique into various adversarial based DA methods~\cite{Ganin16,Zhang2019bridging1,kurmi2019attending} and show its superiority over the state-of-the-art methods in both UDA and SSDA settings. 

\section{Related Works}
\noindent\textbf{Deep Domain Adaptation.}
Domain adaptation aims to transfer knowledge between a labeled source domain and an unlabeled or a less labeled target domain. 
The key challenge of DA is the existence of domain shift~\cite{0DOI}, the data bias between source and target domains.
There have been many distance-based and divergence-based methods proposed in recent years for measuring and resolving the domain shift proposed in DA.
These measurement dimensions include Maximum Mean Discrepancy between the feature embeddings of different domains~\cite{Long16,inproceedings,Jian2,2021Coarse},
the optimal transmission distance across domains~\cite{courty2016optimal},
high-dimensional discriminative adversarial learning~\cite{Ganin16,zhang2021domain,Tzeng17,Luo,hu2021self}, and so on.
Although these methods are capable of aligning distribution between domains,
they largely ignore the deviation between the estimated transferability and the real one.
Our work focuses on learning unbiased transferability by modeling uncertainty variance of a discriminator in adversarial-based domain adaptation.

\noindent\textbf{Uncertainty Estimation.}
Uncertainty estimation can be used to either measure the uncertainty caused by noise (known as aleatoric uncertainty)
or learn the uncertainty of a model (known as epistemic uncertainty)~\cite{der2009aleatory,kendall2017uncertainties}.
In vision tasks, it is often more challenging and practical to model the epistemic uncertainty,
which can be learned by Bayesian neural network~\cite{blundell2015weight,louizos2017multiplicative}.
Besides, some approximate reasoning methods~\cite{2020Transferable,2017Implicit} can be used to model the uncertainty
based on abundant observation data.
Our work employs the MCDropout~\cite{2015Dropout}, an efficient method for acquiring Bayesian uncertainty,
to model uncertainty variance of a discriminator in adversarial-based domain adaptation.

\noindent\textbf{Uncertainty Modeling in Domain Adaptation.}
Uncertainty modeling can be used to perform cross-domain calibration and improve the reliability of pseudo labels in DA.
CPCS~\cite{2020Calibrated} incorporates importance weighting into temperature scaling to cope with the cross-domain calibration.
TransCal~\cite{2020Transferable} uses post-hoc transferable calibration to realize more accurate calibration with lower bias and variance.
PACET~\cite{liang2019exploring} employs uncertainty modeling to describe the cross-domain distribution differences and other intra-domain relationships.
BUM~\cite{wen2019bayesian} uses a Bayesian neural network to quantify prediction uncertainty of a classifier with the aid of a discriminator.
CADA~\cite{kurmi2019attending} utilizes certainty based attention to identify adaptive
region on pixel level with Bayesian classifier and discriminator.
Besides, uncertainty can also be used to facilitate
cross-domain object detection~\cite{guan2021uncertainty} and segmentation~\cite{zheng2021rectifying}.
Existing approaches mostly focus more on modeling the classifier to obtain better calibrated prediction.
However, due to the extremely unbalanced amount of annotated data in domains, the classifier output is strongly biased to the source, while the discriminator is unbiased.
Our work focuses on learning unbiased transferability by uncertainty modeling the discriminator in theory.
%
Table~\ref{tab:compare} compares representative uncertainty methods.
\section{Methodology}
\subsection{Preliminaries}
\label{sec:3-1}
\noindent\textbf{Problem Definition.}
Let $x_i$ denote an input of the network, $y_i$ is the corresponding class label of $x_i$, and $d_i$ is the domain label of $x_i$.
We set $d_i=1$ when $x_i$ is from the source domain and $d_i=0$ when $x_i$ is from the target one.
In unsupervised domain adaptation (UDA), the labeled source domain is defined as $\mathcal{D}_s = \{({x}_i^s,y^s_i)\}_{i=1}^{n_s}$ with $n_s$ source samples
and the unlabeled target domain is defined as $\mathcal{D}_{t}={\mathcal{D}_{tu}} = \{ {{x}}_i^t\} _{i = 1}^{{n_{tu}}}$ with $n_{tu}$ samples (here, $n_t=n_{tu}$).
$\mathcal{D}_s$ and $\mathcal{D}_t$ are sampled from the source $p_{s}(x)$ and the target $p_{t}(x)$ distributions.
In semi-supervised domain adaptation (SSDA), in addition to the labeled source domain $\mathcal{D}_s$ and the unlabeled target domain ${\mathcal{D}_{tu}}$,
there is an extra target labeled domain $\mathcal{D}_{tl} = \{({x}_i^{tl},y_i^{tl})\}_{i=1}^{n_{tl}}$ with $n_{tl}$ samples,
so the whole target domain can be defined as $\mathcal{D}_{t} =\mathcal{D}_{tl} \cup \mathcal{D}_{tu}$ with $n_t=n_{tl}+n_{tu}$. 
In our paper, the source label space equals to the target one, there are $C$ categories in both source and target domains.

\noindent\textbf{Classical Adversarial Based UDA.}
In the classical adversarial based domain adaptation framework, DANN~\cite{Ganin16},
there is a feature extractor and a domain discriminator,
where the discriminator tries to find out which domain the features come from while the feature extractor tries to confuse the discriminator.
When the discriminator is fully confused, we consider the source classifier can generalize well to the target domains.
In this case, the output of the discriminator should be 0.5, \small{$P(d=0|x)=P(d=1|x)=0.5$}.
Overall, the source classifier $G_y$, the domain discriminator $G_d$ and the feature extractor $G_f$ can be jointly learned by:
\begin{align}
\label{eq:ad_original}
     &\notag\tiny{\min_{G_y}\min _ { G_f } \max _ { G_d }}\mathcal{L}_{adv}(G_d,G_f,G_y)
    =\mathbb { E } _ { (x,y)  \sim p _ { s} }[\mathcal{L}_y(G_y (G_f (x )),y )] \\&-\alpha_{adv}\left(\mathbb { E } _ { x  \sim p _ { s} }[\log(G_d (G_f (x ))]
    + \mathbb { E } _ { x  \sim p _ { t} }[\log(1-G_d (G_f (x )))]\right),
\end{align}
where $\mathcal{L}_y$ is the cross-entropy loss for source classifier. $\alpha_{adv}$ is a trade-off between classifier loss $\mathcal{L}_y$ and discriminator loss $\mathcal{L}_d$. By default, we set $\alpha_{adv}=1$.
\subsection{Unbiased Transferability Estimation}
\label{sec:3-2}
\noindent\textbf{Transferability in DA.}
The challenge of DA is to use the source classifier to minimize the target classification error
$\mathbb { E } _ { (x,y)  \sim p _ { t} }[\mathcal{L}_y(G_y (G_f (x )),y )]$ when target labels are unavailable.
Since the existence of domain shift, the source classifier cannot work properly on target samples.
To better account for the shift between source and target distributions, density ratio
\small{$w(x)=\frac{p_{t}(x)}{p_{s}(x)}$}~\cite{2020Transferable,you2019towards}
can be used as a metric of transferability to measure the discrepancy between domains. In general, the shift will be eliminated when $w(x)=\frac{p_{t}(x)}{p_{s}(x)}=1$.
Then, the target classification error can be estimated by the source distribution $p_s(x)$ as:
\begin{small}
\begin{equation}
\begin{split}
\label{eq:la}
    & \mathbb { E } _ { (x,y)  \sim p_ { t} }\left[\mathcal{L}_y(G_y (G_f (x )),y ))\right]
    = \int_{D_t}\mathcal{L}_y\left(G_y (G_f (x)) ,y\right)p_{t}(x)dx\\
    = &  \int_{D_s}\frac{p_{t}(x)}{p_{s}(x)}\mathcal{L}_y(G_y (G_f (x)) ,y)p_s(x)dx
    =  \mathbb { E } _ { (x,y)  \sim p _ { s} }\left[w(x)\mathcal{L}_y(G_y (G_f (x )),y ))\right].
\end{split}
\end{equation}
\end{small}%
However, the density ratio $w(x)$ is often not accessible in DA, we follow~\cite{2020Transferable} and use estimated $\Hat{w}(x)$ to approximate the real $w(x)$.
Specifically, LogReg~\cite{bickel2007dirichlet,qin1998inferences} is used to estimate the density ratio by Bayesian formula:
\begin{footnotesize}
\begin{equation}
\begin{split}
\label{eq:w}
  \hat{w}(x) & = \frac{p_{t}(x)}{p_{s}(x)}=\frac{m(x|d=0)}{m(x|d=1)} =\frac{P(d=1)P(d=0|x)}{P(d=0)P(d=1|x)} = \frac{n_s}{n_t}\cdot\frac{P(d=0|x)}{P(d=1|x)} = \frac{P(d=0|x)}{P(d=1|x)},
\end{split}
\end{equation}
\end{footnotesize}%
where $m$ is a distribution over $(x,d) \sim X \times (0,1)$, and $d\sim Bernoulli(0.5)$ is a Bernoulli variable representing which domain $x$ belongs to.
Here, $\frac{n_s}{n_t}$ is a constant regarding to sample sizes and the source or the target dataset is randomly up-sampled to ensure $n_s = n_t$.
In this way, $\hat{w}(x)$ only depends on {{$\frac{P(d=0|x)}{P(d=1|x)}$}}. 

In DA, ${w}(x)$ can be treated as the real transferability, while $\hat{w}(x)$ can be considered as the estimated transferability.
Ideally, when ${w}(x) = \hat{w}(x)$, domain adaptation is unbiased.
Previous DA methods~\cite{Long15,inproceedings} mainly focus on aligning distributions between source and target domains to encourage $\hat{w}(x)=1$ 
but largely ignore the deviation between the estimated $\hat{w}(x)$ and the real ${w}(x)$, which yields \emph{biased domain adaptation}.

{\noindent\textbf{Learning Unbiased Transferability.}
There have been some DA methods~\cite{2020Transferable,wen2019bayesian} employing the classifier output for uncertainty modeling to encourage transferability/calibration.
However, due to the lack of class labels in the target domain,
using predictions of unlabeled target samples from the source domain classifier
merely describes the alignment of target samples to the source distribution, while not encouraging the source distribution to align to the target one.
On the other hand, domain labels are always known for both source and target samples during model learning,
which provides abundant label information for measuring the bias between the estimated $\hat{w}(x)$ and the real ${w}(x)$.
In light of this, to account for the deviation between the estimated $\hat{w}(x)$ and the real $w(x)$,
\emph{we propose to model the transferability by uncertainty variance estimation of a discriminator in adversarial-based DA.}

Specifically, different from previous classifier-based uncertainty modeling methods, we introduce a novel unbiased transferability estimation method by lowering the variance of the discriminator output \small${{\mathbb{V}ar _ { x  \sim p_{ds}}{(\hat{P_d}(d=1|x))}}}$
and \small${{\mathbb{V}ar _ { x  \sim p_{dt}}{(\hat{P_d}(d=0|x))}}}$.
Here, $p_{ds}$ and $p_{dt}$ are the source and target discriminator distributions, an instance is sampled from $p_d$ which equals to be sampled from $p_{ds}$ or $p_{dt}$. 
We use MCDropout~\cite{2015Dropout,wen2019bayesian} to compute ${\mathbb{V}ar _ {x  \sim                         p_{d}}{(\hat{P_d}(d|x))}}$ as:\par}
\begin{small}
\begin{align}
\label{eq:mcdrop_variance}
\quad\mathbb{V}ar _ {x\sim p_{d}}{(\hat{P_d}(d|x))}
\approx\footnotesize{\frac{1}{K}\sum_{k=1}^K\left((G_d(G_g(x)))_k-\left(\frac{1}{K}\sum_{k=1}^K(G_d(G_g(x)))_k\right)\right)^2},
\end{align}
\end{small}%
where $K$ is the number of times performing stochastic forward passes through the discriminator network.
We set{ \small{$ u(x) = \mathbb{V}ar _ { x  \sim p_{ds} }{(\hat{P_d}(d=1|x))} + \mathbb{V}ar _ { x \sim p_{dt} }{(\hat{P_d}(d=0|x))}$}},
the modeled uncertainty. Here, 
we set $\mathcal U =[u(x_1), ...u(x_i),...]$,
where $u(x_i)$ measures the uncertainty of the $i$th sample.
Since the uncertainty can also be seen as a measurement of the distance between $x$ and the general distribution (initial aligned distribution of source and target samples, which be called $p_{s,t}$ in the following). When a sample $x$ is close to  $p_{s,t}$, $u(x)$ is minimized and $x$ possesses better transferability.
From this perspective, $u(x)$ can also be seen as a transferability metric.
%
We normalize $u(x)$ by Min-Max Normalization as follows:
\begin{small}
\begin{equation}
\begin{split}
\label{eq:normalize}
\mu(x) =\frac{u(x)-\min(\mathcal U)}{\max(\mathcal U)},
\end{split}
\end{equation}
\end{small}%
Then, we set normalized unbiased transferability $\mu(x)$ as the transferability weight in the adversarial learning process (Eq.~\eqref{eq:ad_original})
to facilitate unbiased DA model learning as:
%
\begin{equation}
\begin{split}
\label{eq:weight}
    &\tiny{\min_{G_y}\min _ { G_f } \max _ { G_d }}\mathcal{L}_{adv}(G_d,G_f,G_y) 
    = \mathbb { E } _ { (x,y)  \sim p _ { s} }[\mathcal{L}_y(G_y (G_f (x )),y )] \\
- &  \mathbb { E } _ { x  \sim p _ { s} }[(1+\mu(x))\log(G_d (G_f (x )))]
  - \mathbb { E } _ { x  \sim p _ { t} }[(1+\mu(x))\log((1-G_d (G_f (x )))]. 
\end{split}
\end{equation}
%
\noindent Furthermore, based on Eq.~\eqref{eq:mcdrop_variance}, we define the L2 regularized $\mathcal U$ as the bias loss and minimize the bias loss as:
\begin{small}
\begin{equation}
\begin{split}
\label{eq:bias_loss}
    \min \mathcal{L}_{bias} &= \min{\left|\left|{\mathcal U}\right|\right|_2^2}
    = \min_{G_d,G_f}\sum_{i=1}^{\tiny{n_t+n_s}}{\left({\mathbb{V}ar _ { x_{i}  \sim p_{d}}{(\hat{P_d}(d|x_i))}}\right)^2}.
\end{split}
\end{equation}
\end{small}%

\noindent With each mini-batch, we perform K times stochastic forward passes through
the discriminator to estimate the variance in Eq.~\eqref{eq:bias_loss}.

\noindent \textbf{Theoretical Analysis.}
In this section, we theoretically analyze the cause of the bias in transferability and discuss
how to lower the upper bound of the bias.
With Eqs.~\eqref{eq:la} and~\eqref{eq:w}, we show how to measure the target classification error
and the estimated transferability. 
Following~\cite{2020Transferable}, we can use the difference between the estimated target classification error and the real one
to measure the bias of transferability, as:
%
\begin{align}
\label{eq:ineq}
\notag 
     &  \left|\mathbb { E } _ { (x,y)  \sim p _ { t} }[\mathcal{L}_y^{\hat{w}(x)}(G_y (G_f (x )),y ))]
     -\mathbb { E } _ { (x,y)  \sim p _ { t} }[\mathcal{L}_y^{w(x)}(G_y (G_f (x )),y ))]\right| 
    \\ 
      = & \left|\mathbb { E } _ { (x,y)  \sim p _ {s} }[(\hat{w}(x)-w(x))\mathcal{L}_y(G_y (G_f (x )),y ))]\right|
     \\ \notag
    \le &\frac{1}{2}\left(\mathbb { E } _ { x  \sim p _ {s} }\left[(\hat{w}(x)-w(x)\right)^2\right]+\mathbb { E } _ { (x,y)  \sim p _ {s} }\left[\left(\mathcal{L}_y(G_y (G_f (x )),y ))^2\right]\right). 
\end{align}
In the above inequality, since the second term is bounded by supervised learning in the labeled source domain,
we only need to focus on the first term.
We use a discriminator to alleviate the deviation between the estimated $\hat{w}(x)$ and the real ${w}(x)$.
From our unbiased transferability perspective, we further formalize the transferability based on discriminator as $W(x)=\frac{B_{t}(x|d=0)}{B_{s}(x|d=1)}$.
Here, distribution $B$ is a distribution over $(x,d) \in X (0,1)$. In this case, $d\sim Bernoulli(0.5)$, if $d=1$, $x\sim{p_s}$ or $x\sim{p_t}$.
Furthermore, as the unbiased transferability is derived in the discriminator label space, $W(x)$ and $\hat{W}(x)$ are assumed to be the real and estimated transferability in this space.
Assume we have upper bound {$ N\ge 0 $} for $W(x)$ subject to { $ N \ge W(x) \ge 0 $} according to the bounded importance weight assumption~\cite{cortes2010learning}.
Combined with upper bound $N$, we have $\frac{1}{N+1} \le P_d(d=1|x) \le 1$ and $P_d(x)=B(d=1|x)=\frac{1}{1+W(x)}$.
Then, the first term of Eq.~\eqref{eq:ineq} is bounded by:
\begin{footnotesize}
\begin{align}\notag
\label{eq:ineq_first_term}
     &\notag\mathbb { E } _ { x  \sim p _ {s} }\left[(\hat{w}(x_i)-w(x))^2\right] 
     = \mathbb { E } _ { x  \sim p _ {ds} }\left[(\hat{W}(x)-W(x))^2*\frac{p_s(x)}{p_{ds}(x)}\right] \\
     \le &\notag 2\mathbb { E } _ { x  \sim p _ {ds} }[(\hat{W}(x_i)-W(x))^2] 
     = 2\mathbb { E } _ { x  \sim p _ {ds} }\left[\left(\frac{P_d(d=1|x)-\Hat{P_d}(d=1|x)}{P_d(d=1|x)\Hat{P_d}(d=1|x)}\right)^2\right] \\
     \le& 2(N+1)^4 \mathbb { E } _ { x  \sim p _ {ds}}{\left[\left(P_d(d=1|x)-\Hat{P_d}(d=1|x)\right)^2\right]}.
\end{align}
\end{footnotesize}%
{The first row changes the probability from source label space to source discriminator domain label space. Then, the deviation between the real and the estimated transferability is calculated in the discriminator label space. The second inequality in Eq.~\eqref{eq:ineq_first_term} can be further rewritten as:\par}
\begin{footnotesize}
\begin{align}
\begin{medsize}
\end{medsize}
\label{eq:ineq_our}
     &\notag\quad2(N+1)^4 \mathbb { E } _ { x  \sim p _ {ds}}{\left[\left(P_d(d=1|x)-\Hat{P_d}(d=1|x)\right)^2\right]} \\
     = & \notag 2(N+1)^4 \left(
     \mathbb { E } _ { x  \sim p _ {ds}}{[({P_d}(d=1|x))^2]} -\left(\mathbb { E } _ { x  \sim p _ {ds}}{\left[P_d(d=1|x)\right]}\right)^2 \right. \\
     & \notag \left.+\left(\mathbb { E } _ { x  \sim p _ {ds}}{\left[P_d(d=1|x)\right]}\right)^2 \right. + \mathbb { E } _ { x  \sim p _ {ds}}{[(\Hat{P_d}(d=1|x))^2]}\\\notag
     & \notag \left. 
     -(\mathbb { E } _ { x  \sim p _ {ds}}{[\Hat{P_d}(d=1|x)]})^2+(\mathbb { E } _ { x  \sim p _ {ds}}{[\Hat{P_d}(d=1|x)]})^2 \right. \\
     & \notag \left.- 2\mathbb { E } _ { x  \sim p _ {ds}}{[\Hat{P_d}(d=1|x)P_d(d=1|x)]}\right ) \\\notag
     = & 2(N+1)^4 \left ( \mathbb{V}ar _ { x  \sim p _ {ds}}{\left({P_d}(d=1|x)\right)}+\mathbb{V}ar _ { x  \sim p _ {ds}}{(\Hat{P_d}(d=1|x))}\right. \\
     &+\left.\left(\mathbb { E } _ { x  \sim p _ {ds}}{\left[P_d(d=1|x)\right]}-\mathbb { E } _ { x  \sim p _ {ds}}{[\Hat{P_d}(d=1|x)]}\right)^2 \right),
\end{align}
\end{footnotesize}%
where the variances of outputs under real and estimated probability distributions are $\mathbb{V}ar _ { x  \sim p _ {ds}}{\left({P_d}(d=1|x)\right)}$ and $\mathbb{V}ar _ { x  \sim p _ {ds}}{(\Hat{P_d}(d=1|x))}$  respectively.
Ideally, when the source and target domains are aligned, the output of discriminator should be 0.5,
then, $\mathbb{V}ar _ { x  \sim p _ {ds}}{({P_d}(d=1|x))}=0$.
Hence, the bias of transferability can be formulated as:
\begin{small}
\begin{align}
\label{eq:dpdgan_equ}\notag
    \quad\mathbb { E } _ { x  \sim p _ {ds} }&\Big[( \hat{w}(x)-w(x))^2\Big]   \le2(N+1)^4 \Big [\mathbb{V}ar _ { x  \sim p _ {ds}}{\left(\Hat{P_d}(d=1|x)\right)} \\&   + \left(\mathbb { E } _ { x  \sim p _ {ds}}{\left[P_d(d=1|x)\right]}-\mathbb { E } _ { x  \sim p _ {ds}}{[\Hat{P_d}(d=1|x)]}\right)^2 \Big] .
\end{align}
\end{small}%
The second term of Eq.~\eqref{eq:dpdgan_equ} is constrained since the domain adaptation process encourages   ${\mathbb { E } _ { x  \sim p _ {ds}}{(P_d(d=1|x))}}$ to approximate to
${\mathbb { E } _ { x  \sim p _ {ds}}{(\hat{P_d}(d=1|x))}}$.
Therefore, to learn unbiased transferability, we can minimize ${\mathbb{V}ar _ { x  \sim{p} _ {ds}}{(\hat{P_d}(d=1|x))}}$.
Besides, since we need to use the estimated uncertainty of unlabeled samples from the target domain for pseudo label selection (see Sec.~\ref{sec:3-3}),
we also minimize ${\mathbb{V}ar _ { x  \sim{p} _ {dt}}{(\hat{P_d}(d=0|x))}}$ and use it as a part of the transferability weight.
Thus, we set:
\begin{equation}
   u(x) = \mathbb{V}ar _ { x  \sim p_{ds} }{(\hat{P_d}(d=1|x))} + \mathbb{V}ar _ { x  \sim p_{dt} }{(\hat{P_d}(d=0|x))}.
\end{equation}

In this way, for both the source and target samples, we lower the variance of the discriminator outputs with Eq.~\eqref{eq:mcdrop_variance},
and use Eq.~\eqref{eq:bias_loss} to lower the upper bound of the deviation between estimated transferability and the real one
to realize unbiased domain adaptation. The details of the theoretical analysis are in the supplementary materials.

\subsection{Unbiased Domain Adaptation}
\label{sec:3-3}
%
\noindent\textbf{Pseudo Label Selection.} Originally, pseudo label is introduced to solve the problem of label shortage in unlabeled domain in semi-supervised learning.
However, recent works~\cite{chen2019progressive,liu2021cycle} implies that when domain shift exists, the pseudo label selection strategies tailored for semi-supervised learning is difficult to be effective. 
Different from the classifier-based pseudo label evaluation methods~\cite{2021In,french2018self,yang2021dense}, we evaluate pseudo label reliability under domain shift by the normalized unbiased transferability $\mu(x)$ of domain discriminator.
Intuitively, the lower $\mu(x)$ indicates the better transferability of $x$, which is more reliable for pseudo labeling.
Denote $s(x)$ as the selected weight for $x$:
\begin{equation}
    s(x)=1-\mu(x).
\end{equation}
Then, pseudo labels are generated with those preliminary refined unlabeled samples of the target domain based on the predefined thresholds.
Specifically, suppose $g(x)$ is the $C$-ways source classifier probability prediction output for sample $x$,  $g(x)=G_y(G_f(x))=[g^{[1]}(x),...g^{[c]}(x),...,g^{[C]}(x)]$.
Here, $g^{[c]}(x)$ is the probability of class $c$ for the sample. 
Only when the $g^{[c]}(x)$ is higher than the threshold $\beta$, where $\beta \in (0,1)$, the positive pseudo label for the sample is selected.
Hence, $h(x)=[h^{[1]}(x),...,h^{[C]}(x)]$ is a binary vector representing the selected positive pseudo label for the sample.
When $g^{[c]}(x)$ is selected, $h^{[c]}(x)$ is 1, or $h^{[c]}(x)$ is 0.
Here, $h^{[c]}(x)$ is obtained by:
\begin{equation}
    h^{[c]}(x)=\mathbbm{1}[g^{[c]}(x)\ge\beta].
\end{equation}
Then, the selected $g^{[c]}(x)$ is treated as a soft positive pseudo label, so the positive cross-entropy loss is defined as:
\begin{equation}
\begin{split}
     \mathcal{L}_{pce} \left(g,h\right) =-\mathbb { E } _ { x  \sim p _ {d} }{s(x)} \sum_{c=1}^C h^{[c]}(x) \left[g^{[c]}(x)\cdot\log(g^{[c]}(x))\right].
\end{split}
\end{equation}
%
Similarly, we pick out those unlikely categories with high probability as negative pseudo labels to further dismiss the interference of noises on training. Only when $g^{[c]}(x)\le \gamma$, where $\gamma \in (0,1)$, the negative pseudo label is generated for the sample.
Thus, we have:
\begin{equation}
    l^{[c]}(x)=\mathbbm{1}[g^{[c]}(x)\le \gamma],
\end{equation}
where $l(x)$ is corresponding to $h(x)$, representing the selected negative pseudo label for the sample.
Then, the selected $g^{[c]}(x)$ is seen as a soft negative pseudo-label, so a negative cross-entropy loss is defined as:
\begin{small}
\begin{equation}
\begin{split}
      \mathcal{L}_{nce} (g,l)  =-\mathbb { E } _ { x  \sim p _ {d} }{s(x)}{\sum_{c=1}^C} l^{[c]}(x)\left[(1-g^{[c]}(x))\cdot\log(1-g^{[c]}(x))\right].
    \end{split}
\end{equation}
\end{small}%
Thus, the pseudo label cluster learning loss is modeled as:
\begin{equation}
\begin{split}
\label{eq:cluster_loss}
    \mathcal{L}_{tce} (g,h,l)
    = \mathcal{L}_{pce} (g,h) + \alpha_{nce}\mathcal{L}_{nce} (g,l) .
\end{split}
\end{equation}
By default, $\alpha_{nce}=1$.
Note that this strategy is similar to UPS~\cite{2021In} which presents uncertainty-aware pseudo labeling for semi-supervised learning,
but we conjecture that the uncertainty of labels mostly comes from the domain gap between labeled and unlabeled data.
Hence, unlike UPS, uncertainty is derived by a variance of the discriminator output and used for pseudo label selection to facilitate unbiased domain adaptation.
Such a strategy is not only suitable for DA and SSDA, but also for semi-supervised learning problems. 
Furthermore, we conducted a comparative experiment in Table~\ref{table:accuracy_ov4} to show the effectiveness of our uncertainty-aware pseudo labeling.

\noindent\textbf{Unbiased Domain Adaptation.}
The proposed UTEP for DA includes three parts,
namely unbiased domain alignment with adversarial learning, transferability bias regularization, and pseudo label cluster learning. The overall loss of UTEP is defined as:
\begin{equation}
    \mathcal{L} = \mathcal{L}_{adv}+\alpha_{bias} \mathcal{L}_{bias}+\alpha_{tce} \mathcal{L}_{tce},
\end{equation}
where $\mathcal{L}_{adv}$ focuses on domain alignment with the adversarial loss and the source classification loss.
$\mathcal{L}_{bias}$ is the unbiased transferability estimation loss which is obtained per batch during model training. $\mathcal{L}_{tce}$ is the pseudo label cluster loss. $\alpha_{bias}$ and $\alpha_{tce}$ are the hyper-parameters.

\section{Experiments}
\subsection{Datasets and Protocols}
\noindent{\textbf{Datasets.}}
To evaluate the effectiveness of the proposed UTEP approach,
we conduct extensive experiments on three popular DA datasets: \textbf{Office-31}, \textbf{Office-Home} and \textbf{VisDA-2017}.
\textbf{Office-31} is the most popular DA dataset, containing 31 classes and 4600 images from three domains,
namely Webcam(\textbf{W}), Amazon(\textbf{A}) and Dslr(\textbf{D}).
\textbf{Office-Home} includes 15,500 images from 65 categories collected from four domains,
namely Artistic Images (\textbf{Ar}), Clip Art (\textbf{Cl}), Product (\textbf{Pr}) and Real-World (\textbf{Rw}).
\textbf{VisDA-2017} is a more challenging \textbf{S}imulation-to-\textbf{R}eal dataset with more than 280K images in 12 categories.

\noindent{\textbf{Protocols.}}
Our experiments are performed under two different settings, namely UDA and SSDA,
and are carried out using PyTorch.
Our codes are based on \cite{dalib} and released on \href{https://github.com/puchapu/UTEP}{Github}.
We use mini-batch SGD to fine-tune an ImageNet pretrained model (ResNet-50~\cite{He} and ResNet-101 for different DA tasks)
as the feature encoder with the learning rate as 0.001
and to learn the new layers (bottleneck layer and classification layer) from scratch with the learning rate is 0.01.
In the UDA setting, each training batch consists of 32 source samples and 32 target samples.
In the SSDA setting, each training batch consists of 16 source samples, 16 labeled target samples and 32 unlabeled target samples. 
We conduct the experiment with ResNet-34 as the feature encoder.
It is worth noting that only 1\% target samples are selected out as the labeled target domain for training.
More implementation details are in the supplementary material.

\begin{table*}[tb!]
    \small
    \centering 
    \caption{Classification accuracy of UDA on \emph{Office-31} with ResNet-50 as backbone model. Best in \textbf{bold} and the second best in \textbf{\underline{bold with underline}}.}
  \resizebox{0.65\linewidth}{!}{%
    \begin{tabular}{c|p{25pt}<{\centering}p{25pt}<{\centering}p{25pt}<{\centering}p{25pt}<{\centering}p{25pt}<{\centering}p{25pt}<{\centering}|c}
        \hline
        \centering Method & A$\rightarrow$W & D$\rightarrow$W & W$\rightarrow$D & A$\rightarrow$D & D$\rightarrow$A & W$\rightarrow$A & Avg \\
        \hline
        MinEnt~\cite{grandvalet2005semi} & 89.4 & 97.5 & \textbf{100.0} & 90.7 & 67.1 & 65.0 & 85.0 \\
        ResNet~\cite{He}& 75.8 & 95.5 & 99.0 & 79.3 & 63.6 & 63.8 & 79.5 \\
        GTA~\cite{Ganin16}& 89.5 & 97.9 & 99.8 & 87.7 & 72.8 & 71.4 & 86.5 \\
        CDAN+E~\cite{2017Conditional}& 94.2 & 98.6 & \textbf{100.0} & 94.5 & 72.8 & 72.2 & 88.7 \\
        SAFN~\cite{2018larger}& 90.1 & 98.6 & 99.8 & 90.7 & 73.0 & 70.2 & 87.1\\
        CAN~\cite{kang2019contrastive}& 94.5 & 99.1  & 99.8 & 95.0 & \textbf{\underline{78.0}} & \textbf{\underline{77.0}} &  \textbf{\underline{90.6}} \\
        MCC~\cite{2020Minimum} & 94.0 & 98.5 & \textbf{100.0} & 92.1 & 74.9 & 75.3 & 89.1 \\
        BNM~\cite{2020Towards}& 94.0 & 98.5  & \textbf{100.0} & 92.2 & 74.9 & 75.3 &  89.2 \\
        GSDA~\cite{hu2020unsupervised} & 95.7 & 99.1 &\textbf{100.0} & 94.8 & 73.5 & 74.9 & 89.7 \\
        SRDC~\cite{tang2020unsupervised}& 95.7 & \textbf{\underline{99.2}}  & \textbf{100.0} & \textbf{95.8} & 76.7 & \textbf{77.1} &  \textbf{{90.8}} \\
        VAK~\cite{Hou_2021_CVPR} & 91.8 & 98.7 & \textbf{\underline{99.9}} & 89.9 & 73.9 & 72.0 & 87.7\\
        \hline
        DANN~\cite{Ganin16}& 91.4 & 97.9 & \textbf{100.0} & 83.6 & 73.3 & 70.4 & 86.1 \\
        \rowcolor{gray!20}
        {\centering DANN+UTEP}& 92.7 & 98.5 & \textbf{100.0} & 90.4 & 73.8 & 72.7 & 88.0\footnotesize{(\textbf{{\color[rgb]{1,0,0}\small{+1.9\%}}})}\\\hline
        MDD~\cite{Zhang2019bridging1}& {94.5} & 98.4 & \textbf{100.0} & 93.4 & 74.6 & 72.2 & 88.9 \\\rowcolor{gray!20}
        MDD+UTEP& 94.7 & 99.0 & \textbf{100.0} & {94.4}  & 77.0 & 74.5& 89.9\footnotesize{(\textbf{{\color[rgb]{1,0,0}\small{+1.0\%}}})}\\\hline
        CADA~\cite{kurmi2019attending}& \textbf{\underline{97.0}} & 99.3 & \textbf{100.0} & \textbf{\underline{95.6}} & 71.5 & 73.0 & 89.5 \\\rowcolor{gray!20}
        \centering CADA+UTEP& \textbf{97.2} & \textbf{99.4} & \textbf{100.0} & 95.3  & 73.5 & 74.3& 90.0\footnotesize{(\textbf{{\color[rgb]{1,0,0}\small{+0.5\%}}})}\\\hline
        TransPar~\cite{han2021learning}& 95.5 & 98.9 & \textbf{100.0} & 94.2 & 77.7 & 72.8 & 89.9 \\\rowcolor{gray!20}
        \centering TransPar+UTEP& 95.7 & \textbf{99.4} & \textbf{100.0} & 95.2 & \textbf{78.6} & 75.6& \textbf{90.8}\footnotesize{(\textbf{{\color[rgb]{1,0,0}\small{+0.9\%}}})}\\
        \hline
    \end{tabular}%
}

\label{table:accuracy_ovw}
\end{table*}

\begin{table*}[tb!]
    \large
    
    \centering 
   \caption{Accuracy of UDA on \emph{Office-Home} with ResNet-50.}
  \resizebox{1.0\textwidth}{!}{%
  \renewcommand\tabcolsep{1.0pt}
    \begin{tabular}{p{90pt}<{\centering}|p{27pt}<{\centering}p{29pt}<{\centering}p{32pt}<{\centering}p{32pt}<{\centering}p{30pt}<{\centering}p{30pt}<{\centering}p{33pt}<{\centering}p{32pt}<{\centering}p{32pt}<{\centering}p{32pt}<{\centering}p{32pt}<{\centering}p{32pt}<{\centering}|p{60pt}<{\centering}}
        \hline
        Method
        & \small{Ar$\rightarrow$Cl} & \small{Ar$\rightarrow$Pr} & \small{Ar$\rightarrow$Rw} & \small{Cl$\rightarrow$Ar} & \small{Cl$\rightarrow$Pr} & \small{Cl$\rightarrow$Rw} & \small{Pr
      $\rightarrow$Ar} &\small{Pr
      $\rightarrow$Cl} & \small{Pr$\rightarrow$Rw} & \small{Rw$\rightarrow$Ar} &\small{Rw$\rightarrow$Cl} & \small{Rw$\rightarrow$Pr} & \small{Avg} \\
        \hline
        {MinEnt~\cite{grandvalet2005semi}}& 51.0 & 71.9 & 77.1 & 61.2 & 69.1 & 70.1 & 59.3 & 48.7 & 77.0 & 70.4 & 53.0 & 81.0 & 65.8\\
        {ResNet~\cite{He}}& 41.1 & 65.9 & 73.7 & 53.1 & 60.1 & 63.3 & 52.2 & 36.7 & 71.8 & 64.8 & 42.6 & 75.2& 58.4\\
        {CDAN+E~\cite{2017Conditional}}&54.6 & 74.1 & 78.1 & 63.0 & 72.2 & 74.1 & 61.6 & 52.3 & 79.1 & 72.3 & 57.3 & 82.8 & 68.5\\
        {DAN~\cite{Zhang18}}& 45.6 & 67.7 & 73.9 & 57.7 & 63.8 & 66.0 & 54.9 & 40.0 & 74.5 & 66.2 & 49.1 & 77.9 & 61.4\\
        {SAFN~\cite{2018larger}}&52.0 & 71.7 & 76.3 & 64.2 & 69.9 & 71.9 & 63.7 & 51.4 & 77.1 & 70.9 & 57.1 & 81.5 & 67.3\\
        {TransCal~\cite{2020Transferable}} & 49.4 & 68.4 & 75.5 & 57.6 & 70.1 & 70.4 &51.1 &
        50.3 & 72.4 & 68.9 &  54.4 & 81.2 &
        64.1 \\
        {ATM~\cite{2020Maximum}}&52.4 &72.6 &78.0 &61.1 &72.0& 72.6 &59.5 &52.0 &79.1 &73.3 &58.9& 83.4 &67.9\\
        {CKD~\cite{Luo_2021_CVPR}} & 54.2 & 74.1 & 77.5 & 64.6 & 72.2 & 71.0 & 64.5 & 53.4 & 78.7 & 72.6 & 58.4 & 82.8 & 68.7 \\
        {GSDA~\cite{hu2020unsupervisedsync}} & \textbf{61.3} & 76.1 & 79.4 & 65.4 & 73.3 & 74.3 & 65.0 & 53.2 & 80.0 & 72.2& 60.6 & 83.1 & 70.3\\
        {SRDC~\cite{tang2020unsupervised}} & 52.3 & 76.3 & \textbf{\underline{81.0}} & \textbf{\underline{69.5}} & \textbf{76.2} & \textbf{78.0} & \textbf{68.7} & 53.8 & \textbf{81.7} & 76.3 & 57.1 & \textbf{85.0} & \textbf{71.3}\\
        {DCC~\cite{Li_2021_CVPR}} & \textbf{\underline{58.0}} & 54.1 & 58.0 & \textbf{74.6} & 70.6 & \textbf{\underline{77.5}} & 64.3 & \textbf{73.6} & 74.9 & \textbf{80.9} & \textbf{75.1} & 80.4 & 70.2 \\
        \hline
        {DANN~\cite{Ganin16}}&53.8 & 62.6 & 74.0 & 55.8 & 67.3 & 67.3 & 55.8 & 55.1 & 77.9 & 71.1 & 60.7 & 81.1 & 65.2 \\\rowcolor{gray!20}
          \centering DANN+UTEP & 50.7 & 68.9 & 77.1 & 58.7 & 72.3 & 71.7 & 59.3 &
        52.9 &
        79.8 &
        73.5 &
        59.9 & 83.6 &
        67.3\footnotesize{(\textbf{{\color[rgb]{1,0,0}\small{+2.1\%}}})}
        \\ 
        \hline
         {MDD~\cite{Zhang2019bridging1}}&54.9 & 73.7 & 77.8 & 60.0  & 71.4 & 71.8 & 61.2 & 53.6 & 78.1 & 72.5 & 60.2 & 82.3 & 68.1 \\\rowcolor{gray!20}
          \centering MDD+UTEP & 57.2 & 75.9 & 79.6 & 63.4 & 72.8 & 73.7 & 64.6 & \textbf{\underline{55.4}} & 79.8 &
        74.0 &  61.1 &
        84.2 &
        70.1\footnotesize{(\textbf{{\color[rgb]{1,0,0}\small{+2.0\%}}})}
        \\ 
        \hline
        {CADA~\cite{kurmi2019attending}} & 56.9 & \textbf{\underline{76.4}} & 80.7 & 61.3 & 75.2 & 75.2 & 63.2 & 54.5 & 80.7 & 73.9 & 61.5 & 84.1 & 70.2 \\ 
        \rowcolor{gray!20}
          \centering{CADA+UTEP} &57.1 &\textbf{76.5} &\textbf{81.1} &61.7 &\textbf{\underline{75.4}} &75.2 &63.9 &54.9 &\textbf{\underline{80.9}}& 74.2 &\textbf{\underline{61.8}} &84.1 & \textbf{\underline{70.6}}\footnotesize{(\textbf{{\color[rgb]{1,0,0}\small{+0.4\%}}})}\\\hline
         {TransPar~\cite{han2021learning}}& 55.3 & 75.9 & 79.2 & 63.2  & 72.4 & 71.8 & 61.2 & 55.1 & 80.0 & 74.5 & 60.9 & 83.7 & 69.8 \\\rowcolor{gray!20}
          \centering{TransPar+UTEP} & 57.4 & 76.1 & 80.2 & 64.2 & 73.2 & 73.7 & \textbf{\underline{64.8}} & \textbf{\underline{55.4}} & \textbf{\underline{80.9}} &
        \textbf{\underline{74.7}} &  61.1 &
        \textbf{\underline{84.6}} &
        \textbf{\underline{70.6}}\footnotesize{(\textbf{{\color[rgb]{1,0,0}\small{+0.8\%}}})}\\
        \hline
    \end{tabular}%
}

\label{table:accuracy_co1}
\end{table*}

\begin{table*}[tb!]
    \small
    \centering 
    \caption{Classification accuracy of UDA on \emph{VisDA-2107} using ResNet-101.}
  \resizebox{1.0\textwidth}{!}{
  \renewcommand\tabcolsep{1.0pt}
    \begin{tabular}{p{70pt}<{\centering}|p{24pt}<{\centering}p{24pt}<{\centering}p{18pt}<{\centering}p{24pt}<{\centering}p{22pt}<{\centering}p{24pt}<{\centering}p{24pt}<{\centering}p{24pt}<{\centering}p{24pt}<{\centering}p{24pt}<{\centering}p{24pt}<{\centering}p{22pt}<{\centering}|p{50pt}<{\centering}}
        \hline
        Method
        & plane & bicycle& bus & car & horse & knife & mcycl & person & plant & sktbrd & train & truck & Mean \\
        \hline
        MinEnt~\cite{grandvalet2005semi} & 88.6 & 29.5 & 82.5 & 75.8 & 88.7 & 16.0 & 93.2  & 63.4 & 94.2 & 40.1 & 87.3 & 12.1 & 64.3\\
        ResNet~\cite{He}& 55.1 & 53.3 & 61.9 & 59.1 & 80.6 & 17.9 & 79.7 & 31.2 & 81.0 & 26.5 & 73.5 & 8.5 & 52.4\\
        SAFN~\cite{2018larger}& 94.2 & 56.2 & 81.3 & \textbf{\underline{69.8}} & 93.0 & 81.0 & \textbf{\underline{93.0}} & 74.1 & 91.7 & 55.0 & \textbf{90.6} &18.1 & 75.0\\
        MixMatch~\cite{2019MixMatch}&93.9& 71.8 &93.5& \textbf{\underline{82.1}} &\textbf{95.3} &0.7 &90.8 &38.1 & 94.2 &96.0 &86.3 &2.2 &70.4\\
        BNM~\cite{2020Towards}& 91.1& 69.0& 76.7& 64.3& 89.8& 61.2& 90.8& 74.8& 90.9& 66.6& 88.1& 46.1& 75.8\\
        MCC~\cite{2020Minimum} & 92.2 & \textbf{82.9} & 76.8 & 66.6 & 90.9 & 78.5 & 87.9 & 73.8 & 90.1 & 76.1 & 87.1 & 41.0 & 78.7\\
        DWL~\cite{Xiao_2021_CVPR} & 90.7 & \textbf{\underline{80.2}} & 86.1 & 67.6 & 92.4 & 81.5 & 86.8 & 78.0 & 90.6 & 57.1 & 85.6 & 28.7 & 77.1\\
        VAK~\cite{Hou_2021_CVPR} & \textbf{\underline{94.3}} & 79.0 & \textbf{\underline{84.9}} & 63.6 & 92.6 & 92.0 & 88.4 & 79.1 & 92.2 & 79.8 & 87.6 & 43.0 & 81.4 \\
        \hline
        DANN~\cite{Ganin16}& 90.0 & 58.9 & 76.9 & 56.1 & 80.3 & 60.9 & 89.1 & 72.5 & 84.3 & 73.8 & \textbf{\underline{89.3}} & 35.8 & 72.3 \\\rowcolor{gray!20}
          \centering DANN+UTEP & 93.8 & 70.7 & \textbf{\underline{85.5}}    & 62.7 & 91.0   & 90.7  & 88.6  & 76.2  & 87.2  & 85.4  & 84.7 &    33.8 & 79.2\footnotesize{(\textbf{{\color[rgb]{1,0,0}+6.9\%}})}
          \\ \hline
        MDD~\cite{Zhang2019bridging1} & 94.2  & 71.6  & 84.3  & 65.4  & 91.5  & 94.9 & 92.0  & \textbf{\underline{80.3}}  & 90.8  & \textbf{90.0}& 82.4 & 42.0 & 81.4\\
        \rowcolor{gray!20}
        \centering MDD+UTEP& \textbf{94.7}& 75.4    & 83.2  & 60.1  & \textbf{\underline{93.7}} &   95.3 & \textbf{93.1}   & \textbf{82.6} & \textbf{\underline{94.3}} & \textbf{\underline{89.8}} & 84.6 & 41.1 & \textbf{82.3}\footnotesize{(\textbf{{\color[rgb]{1,0,0}+0.9\%}})}
        \\\hline
          CADA~\cite{kurmi2019attending} & 85.1 & 61.2 & 84.4 & 69.0 & 89.8 & \textbf{97.1} & 93.0 & 75.8 & 90.2 & 84.1 & 81.4 & 45.4 & 79.7\\
        \rowcolor{gray!20}
        \centering CADA+UTEP& 88.9 & 76.9 & 82.6 & 65.6 & 90.8 & \textbf{\underline{96.9}} & 90.6 & 78.5 & 86.3 & 86.4 & 83.8 & \textbf{48.8} &  81.3\footnotesize{(\textbf{{\color[rgb]{1,0,0}+1.6\%}})}
        \\\hline
        TransPar~\cite{han2021learning} & 84.2  & 57.7  & 85.3  & 62.7  & 90.0  & 92.9 & 92.3  & 73.9  & \textbf{95.9}  & 86.9& 84.2 & 44.7 & 79.4\\
        \rowcolor{gray!20}
        \centering TransPar+UTEP& 90.0& 74.8    & 82.6  & 66.2  & 91.1 &   95.8 & 91.3   & 77.5 & 89.0 & 88.3 & 82.6 & \textbf{\underline{47.2}} & \textbf{\underline{81.5}}\footnotesize{(\textbf{{\color[rgb]{1,0,0}+2.1\%}})}
        \\\hline
    \end{tabular}%
}

\label{table:accuracy_co1_visda}
\end{table*}

\subsection{Experimental Results}
\noindent{\textbf{Unsupervised DA.}}
To verify the universality of our UTEP, we incorporate it into various adversarial-based UDA methods,
including three classical methods (DANN~\cite{Ganin16}, MDD~\cite{Zhang2019bridging1} and CADA~\cite{kurmi2019attending})
and a state-of-the-art method (TransPar~\cite{han2021learning}).
Tables~\ref{table:accuracy_ovw}, \ref{table:accuracy_co1} and~\ref{table:accuracy_co1_visda} show experimental results on the small-sized Office-31, the medium-sized Office-Home and the more challenging VisDA-2017, respectively.
Overall, our UTEP module can improve the performance of various adversarial-based baseline methods, achieving state-of-the-art performance among UDA methods.
On Office-31, for average accuracy,
UTEP improves DANN (by 1.9\%), MDD (by 1.0\%), CADA (by 0.5\%) and TransPar (by 0.9\%) 
and TransPar+UTEP achieves 90.8\% which is on par with the state-of-the-art.
On Office-Home, in terms of average accuracy,
UTEP significantly improves DANN, MDD and TransPar by approximately 2.0\% and improves CADA by 0.4\%,
while CADA+UTEP and TransPar+UTEP achieve 70.6\% which are the second best results. On VisDA-2017, we can see a notable improvement of UTEP to DANN (by 6.9\%) in average accuracy
while the improvement of UTEP to MDD, CADA and TransPar are also significant, yielding state-of-the-art performance.
In our analysis, the improvement of UTEP to various adversarial-based DA methods can be attributed to learning unbiased transferability by uncertainty variance modeling and uncertainty-based pseudo label selection.
%

\begin{table}[tb!]
  \caption{Classification accuracy of SSDA with 1\% target labeled data on \emph{Office-Home} with ResNet-34 as backbone model (left) and accuracy of 3-shot SSL on \emph{Office-Home} with ResNet-50 as backbone (right).}
    \centering 
  \resizebox{1.0\textwidth}{!}{%
    \begin{tabular}{p{70pt}<{\centering}|ccc|c|p{70pt}<{\centering}|cccc|c}
        \hline
        {\multirow{2}{*}{Methods}} & \multicolumn{4}{c|}{SSDA setting} &    {\multirow{2}{*}{Methods}} & \multicolumn{5}{c}{  3-shot SSL setting} \\ \cline{2-5} \cline{7-11}
        &  Ar$\rightarrow$Rw  &  Rw$\rightarrow$Pr & Pr$\rightarrow$Cl   & Avg  &  
        & Ar & Cl & Pr & Rw & Avg  \\
        \hline
        ADR~\cite{2017Understanding}&70.6 &76.6 &49.5 & 65.6 & ResNet~\cite{He}&48.7 &42.1 &68.9& 66.6 &56.6\\
        IRM ~\cite{2019invariant}&71.1 &77.6 &51.5 &66.7 &MixMatch~\cite{2019MixMatch}&52.2 &41.9 &73.1 &69.1 &59.1\\
        MME~\cite{2019Semi}&72.1 & 78.1 & 52.8  &67.7 & MinEnt ~\cite{grandvalet2005semi}&51.7& 44.5 &72.4 &68.9 &59.4 \\
        CDAN~\cite{2017Conditional}& 73.0 &79.2 &53.1& 68.4 & MCC~\cite{2020Minimum}&58.9 &\textbf{47.7}& 77.4 &74.3 &\textbf{\underline{64.6}}\\
        LIRR~\cite{2020Learning}& \textbf{\underline{73.6}} &\textbf{\underline{80.2}} &\textbf{53.8} & \textbf{\underline{69.2}} & BNM~\cite{2020Towards}& 59.0 &\textbf{\underline{46.0}} &\textbf{\underline{76.5}}& 71.5 &63.2\\
        \hline 
        {\centering DANN~\cite{Ganin16}}& 72.2 & 78.1 & 52.5 &   67.6 & {\centering DANN~\cite{Ganin16}}& 57.4 & 41.6 & 74.5 & 73.4 &  61.7\\
        { DANN+UPS~\cite{2021In}}& 73.2  &  79.6  &  52.8 & 68.5 & { DANN+UPS~\cite{2021In}}&  \textbf{\underline{59.1}}&    42.6 & 75.3 & \textbf{\underline{76.2}} & 63.3\\ 
        \rowcolor{gray!20}
        { DANN+UTEP}&  \textbf{74.4}&   \textbf{82.1} & \textbf{\underline{53.3}}   & \textbf{69.9} \footnotesize{(\textbf{{\color[rgb]{1,0,0}\small{+2.3\%}}})} & { DANN+UTEP}&  \textbf{60.2}&   43.6 & \textbf{77.9}    & \textbf{78.8} & \textbf{65.1}\footnotesize{(\textbf{{\color[rgb]{1,0,0}\small{+3.4\%}})}} \\ 
        \hline
    \end{tabular}%
    }

\label{table:accuracy_ov4}
\end{table}
        


\noindent\textbf{Semi-supervised DA.}
On SSDA, we follow the setting of the state-of-the-art LIRR~\cite{2020Learning}
and conduct experiments on Office-Home.
We randomly select 1\% of target samples as the labeled target domain for training and evaluate SSDA methods on unlabeled samples from the target domain. 
As shown in the left of Table~\ref{table:accuracy_ov4}, our method significantly improves the mean accuracy of DANN (by 2.3\%)
and achieves compelling performance against the state-of-the-art SSDA methods. 
This also examines the efficacy of UTEP for DA under the semi-supervised learning scenario.


\begin{figure*}[tb!]
\flushleft
\subcaptionbox{ResNet\label{fig:ResNet}}{
\begin{minipage}[t]{0.227\linewidth}
\centering
\includegraphics[height=1.28in]{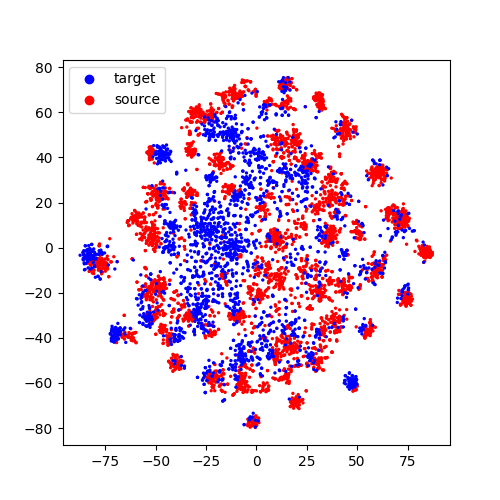}
\end{minipage}%
}%
\subcaptionbox{DANN\label{fig:DANN}}{
\begin{minipage}[t]{0.235\linewidth}
\centering
\includegraphics[height=1.28in]{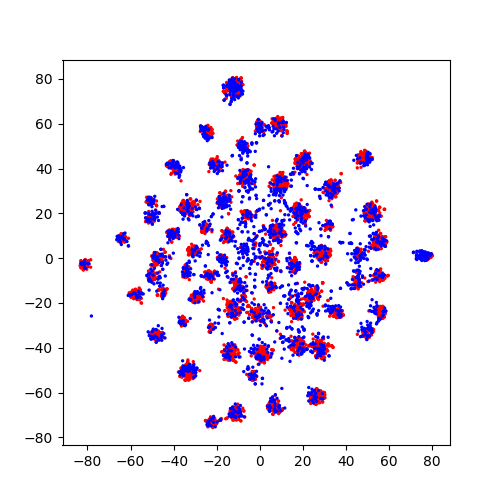}
\end{minipage}
}%
\subcaptionbox{DANN+UPS\label{fig:DANN+UPS}}{
\begin{minipage}[t]{0.21\linewidth}
\centering
\includegraphics[height=1.15in]{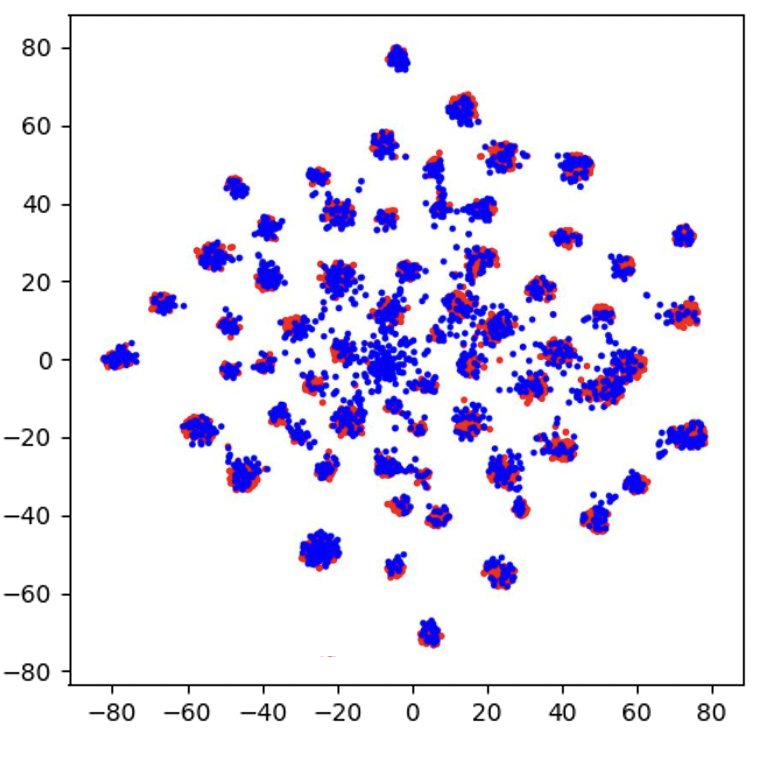}
\end{minipage}
}%
\subcaptionbox{DANN+UTEP\label{fig:DANN_our}}{
\begin{minipage}[t]{0.21\linewidth}
\centering
\includegraphics[height=1.28in]{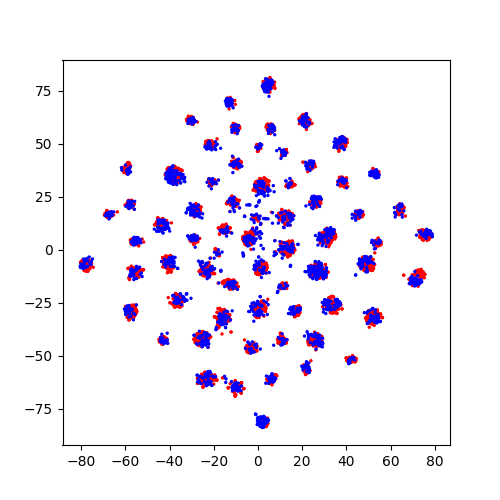}
\end{minipage}
}%
\caption{Visualization results for baseline and ours of UDA w/ ResNet-50.}
\label{fig:analysis}
\end{figure*}
\begin{figure*}[tb!]
\flushleft
\subcaptionbox{Convergence\label{fig:convergence}}{
\begin{minipage}[t]{0.5\linewidth}
\centering
\includegraphics[height=1.6in]{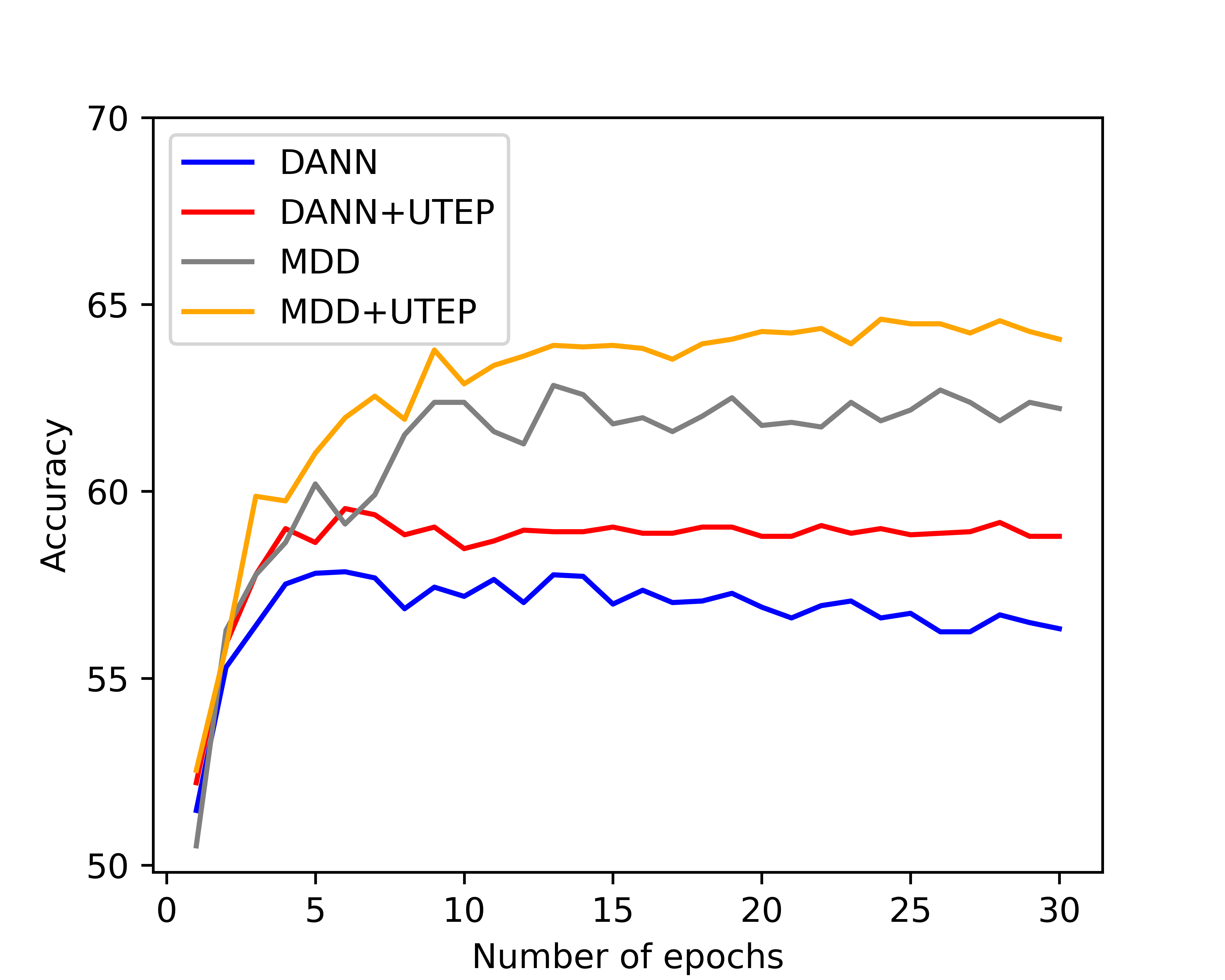}
\end{minipage}%
}%
\subcaptionbox{A-distance\label{fig:adistance}}{
\begin{minipage}[t]{0.5\linewidth}
\centering
\includegraphics[height=1.6in]{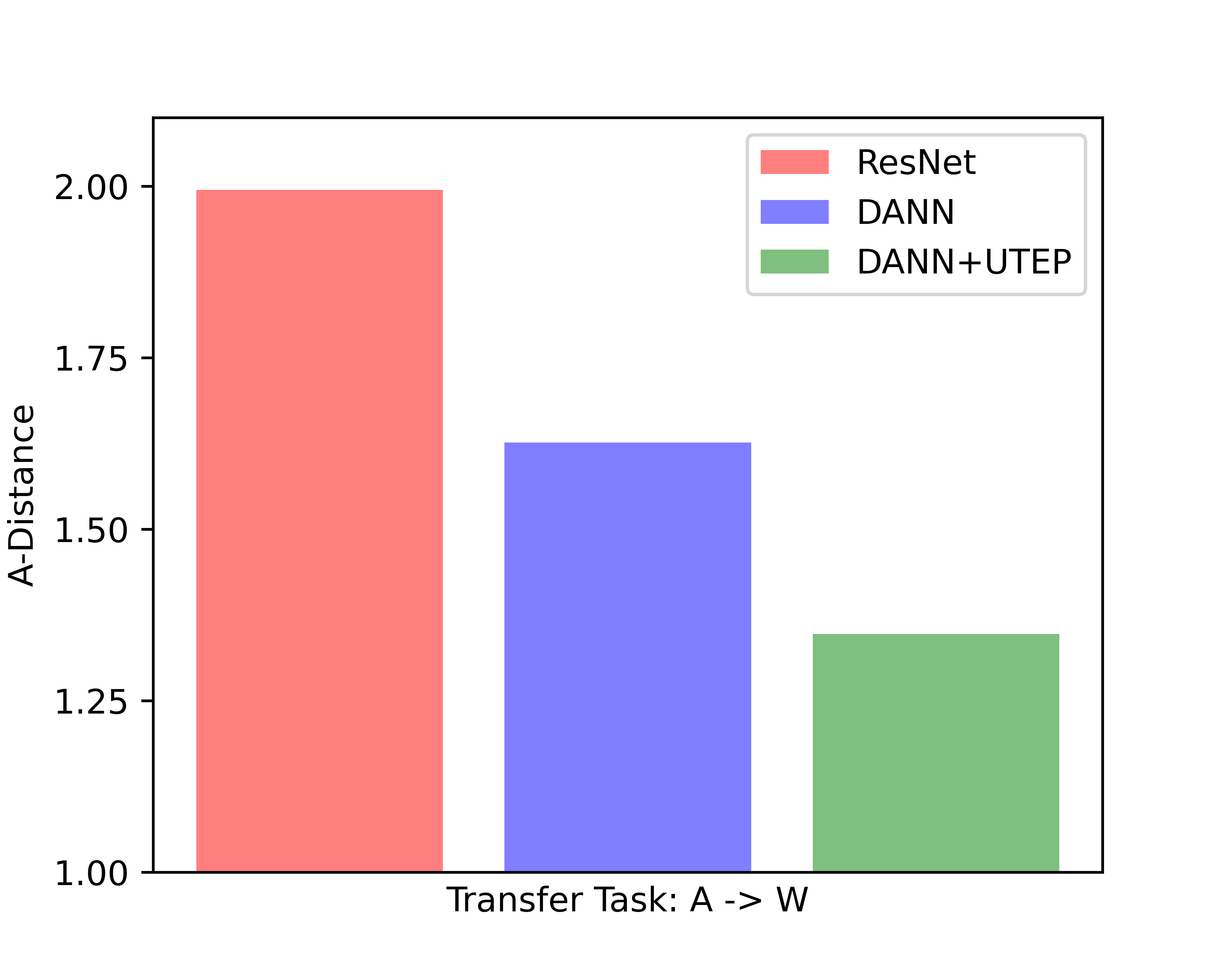}
\end{minipage}%
}%
\caption{(a) Comparison of the training convergence between the two baseline UDA methods(DANN and MDD) and variants equipped with our UTEP with ResNet-50 as backbone; (b) Comparison of the distribution discrepancy measured by A-distance between vanilla ResNet-50, DANN and DANN with our UTEP.}
\label{fig:analysis1}
\end{figure*}
\subsection{Further Analysis and Discussion}
\noindent\textbf{Visualization.}
In Figure~\ref{fig:analysis}(\subref{fig:ResNet})-(\subref{fig:DANN_our}),
we use t-SNE~\cite{donahue2014decaf} to visualize feature embeddings of vanilla ResNet, DANN, DANN+UPS and DANN+UTEP
on task \textbf{Pr}$\rightarrow$\textbf{Ar} (65 classes) on Office-Home.
From Figures~\ref{fig:analysis}(\subref{fig:ResNet}) to \ref{fig:analysis}(\subref{fig:DANN+UPS}), we can see that 
the vanilla ResNet has large domain discrepancy, while DANN significantly decreases the domain discrepancy.
However, the decision boundaries are still not clear in DANN due to the misalignment between the source and target distributions, while DANN+UPS can make the target decision boundary much clearer, but there are still many samples near the boundary are hard to distinguish.
By contrast, as shown in Figure~\ref{fig:analysis}(\subref{fig:DANN_our}), DANN+UTEP further resolves this issue and achieves both better marginal and conditional distribution alignments thanks to the effective learning of unbiased transferability.

\noindent\textbf{Convergence and Distribution Discrepancy.}
In Figure~\ref{fig:analysis1}(\subref{fig:convergence}), we compare the convergence of DANN, DANN+UTEP, MDD and MDD+UTEP
on task \textbf{Pr}$\rightarrow$\textbf{Ar} on Office-Home.
We can see that our UTEP can significantly improve the accuracy of DANN and MDD while achieving more stable convergence.
Figure~\ref{fig:analysis1}(\subref{fig:adistance}) compares A-distance, a widely used measure for distribution discrepancy in DA\cite{ben2006analysis},
of vanilla ResNet, DANN and DANN+UTEP on task \textbf{Pr}$\rightarrow$\textbf{Ar} on Office-Home.
It is clearly shown that A-distance of DANN+UTEP on task \textbf{A}$\rightarrow$\textbf{W}
is smaller than that of vanilla ResNet and DANN, indicating better distribution alignment and higher accuracies of DANN+UTEP.


\begin{table}[tb!]
    \centering 
        \caption{Ablation study of variants with baseline DANN~\cite{Ganin16}. SIW and TIW denote using importance weights for samples from source and target domains respectively (Eq.~\eqref{eq:weight});
    SBL and TBL represent using the bias loss for samples from source and target domains  (Eq.~\eqref{eq:bias_loss});
    PCE and NCE indicate using the positive and negative cross-entropy in pseudo label cluster learning (Eq.~\eqref{eq:cluster_loss}).}
  \resizebox{0.8\textwidth}{!}{%
    \begin{tabular}{cccccc|cccccc|c}
        \hline
        \multicolumn{6}{c|}{\centering DANN~\cite{Ganin16}'s variant} & \multicolumn{7}{|c}{settings on Office-31}\\ \hline
       SIW&TIW&SBL&TBL&PCE&NCE& A$\rightarrow$W & D$\rightarrow$W & W$\rightarrow$D & A$\rightarrow$D & D$\rightarrow$A & W$\rightarrow$A & Avg \\
        \hline
   
        \small\XSolidBrush&\small\XSolidBrush&\small\XSolidBrush&\small\XSolidBrush&\small\XSolidBrush&\small\XSolidBrush&91.4 & 97.9 & \textbf{100.0} & 83.6 & 73.3 & 70.4 & 86.1 \\
        \checkmark&\checkmark&\checkmark&\checkmark&\small\XSolidBrush&\small\XSolidBrush & 92.0 & 98.2 & \textbf{100.0} & 85.7 & 73.0 & 71.1 & 86.7 \\
        \checkmark&\checkmark&\small\XSolidBrush&\XSolidBrush&\checkmark&\checkmark & 92.2 & 98.2 & \textbf{100.0} & 88.5 & 72.9 & 71.5 & 87.2 \\
        \checkmark&\checkmark&\checkmark&\checkmark&\checkmark&\small\XSolidBrush & 92.3 & 98.3 & \textbf{100.0} & 87.9 & 73.3 & 72.0 & 87.3 \\
        \checkmark&\small\XSolidBrush&\checkmark&\small\XSolidBrush&\checkmark&\checkmark & 91.8 & 98.2 & \textbf{100.0} & 88.2 & 73.5 & 70.9 & 87.1\\
        \checkmark&\checkmark&\checkmark&\small\XSolidBrush&\checkmark&\checkmark & 92.0 & 98.2 &\textbf{100.0}  & 89.0 & 73.3 & 71.5 & 87.3\\
        \checkmark&\checkmark&\checkmark&\checkmark&\checkmark&\checkmark & \textbf{92.7} & \textbf{98.5} & \textbf{100.0} & \textbf{90.4} & \textbf{73.8} & \textbf{72.7} & \textbf{88.0}\\
        \hline
    \end{tabular}%
    }

    \label{table:accuracy_ov6}
\end{table}

\noindent\textbf{Component Effectiveness Evaluation.}
In Table~\ref{table:accuracy_ov6}, we study the effectiveness of each component in the proposed UTEP using DANN as the baseline method.
In Table~\ref{table:accuracy_ov6}, the results in the first and seventh rows refer to DANN and DANN+UTEP, respectively.
It shows that all the variants perform better than DANN only, showing the effectiveness of each component in the proposed UTEP.
The results in the second row are those of DANN using uncertainty modeling for sample weighting and bias loss regularization, which are superior to those of DANN only but inferior to those of DANN+UTEP.
The results in the third row shows that without the bias loss, the performance of DANN+UTEP decreases.
The results in the fourth, fifth and sixth rows further show that using parts of the proposed components will also lead to performance degradation.


\noindent\textbf{Evaluation on Semi-Supervised Learning.}
To further study the efficacy of the proposed UTEP, especially compared with semi-supervised pseudo labeling methods (such as UPS~\cite{2021In}),
we conduct 3-shot semi-supervised learning (SSL) experiments on Office-Home in the right part of Table~\ref{table:accuracy_ov4}.
In the SSL setting, only three samples of each class are selected as the labeled domain while the rest are used as the unlabeled domain.
For fair comparison between UTEP and UPS, we incorporate both UTEP and UPS into DANN for experiments.
From the right part of Table~\ref{table:accuracy_ov4}, we can see that both UTEP and UPS can improve the performance of DANN,
which shows the effectiveness of uncertainty-aware pseudo label selection.
Here, DANN+UTEP outperforms DANN+UPS, which further shows the effectiveness of learning unbiased transferability
between labeled and unlabeled data for SSL.

\section{Conclusion}
In this work, we present the Unbiased Transferability Estimation Plug-in (UTEP) to learn unbiased transferability in domain adaptation.
The key idea is to model the variance uncertainty of a discriminator in adversarial-based DA method
and further exploit the uncertainty for pseudo label selection to achieve better marginal and conditional distribution alignment.
Experiments on DA benchmarks show its effectiveness for improving 
various DA methods with state-of-the-art performance.

\noindent\textbf{Acknowledgement.} This work was in part supported by Vision Semantics Limited, Alan Turing Institute, Open Research Projects of Zhejiang Lab (No. 2021KB0AB04), Zhejiang Provincial Natural Science Foundation of China (No. LQ21F020004), and China Scholarship Council.

%
%


\clearpage
%
%
\typeout{}
\bibliographystyle{splncs04}

\end{document}